\PassOptionsToPackage{dvipsnames,svgnames,table}{xcolor}

\documentclass[10pt, a4paper]{article}

\usepackage[]{lrec-coling2024} % this is the new style

\usepackage{latexsym}
\usepackage{graphicx}
\usepackage{graphics}
\usepackage{graphicx}
\usepackage{tabularx}
\usepackage{soul}

\usepackage{microtype}

\usepackage{booktabs} % for \midrule

\newcommand{\comment}[1]{}
\usepackage{pifont}

\newcommand{\cmark}{\ding{51}}%
\newcommand{\xmark}{\ding{55}}%

\usepackage[russian,french,english]{babel}
\usepackage[utf8]{inputenc}
\usepackage[T2A,OT1]{fontenc}
\newcommand\textcyr[1]{{\fontencoding{T2A}\selectfont #1}}

\usepackage{xcolor}
\usepackage{hyperref}
 \definecolor{darkblue}{rgb}{0, 0, 0.5}
  \hypersetup{colorlinks=true, citecolor=darkblue, linkcolor=darkblue, urlcolor=darkblue}

\usepackage{xstring}

\usepackage{color}

\title{GPT-3.5 for Grammatical Error Correction}

\name{Anisia Katinskaia,$^{\star\diamondsuit}$ Roman Yangarber$^{\diamondsuit}$}
\address{ {\bf \large $^{\star}$} Department of Computer Science, {\bf \large $^{\diamondsuit}$} Department of Digital Humanities \\ University of Helsinki, Finland \\ first.last@helsinki.fi}

\abstract{
This paper investigates the application of GPT-3.5 for Grammatical Error Correction (GEC) in multiple languages in several settings: zero-shot GEC, fine-tuning for GEC, and using GPT-3.5 to re-rank correction hypotheses generated by other GEC models.
In the zero-shot setting, we conduct {\em automatic} evaluations of the corrections proposed by GPT-3.5 using several methods: estimating grammaticality with language models (LMs), the Scribendi test, and comparing the semantic embeddings of sentences.
GPT-3.5 has a known tendency to over-correct erroneous sentences and propose alternative corrections.
For several languages, such as  Czech, German, Russian, Spanish, and Ukrainian, GPT-3.5 substantially alters the source sentences, including their semantics, which presents significant challenges for evaluation with reference-based metrics.
For English, GPT-3.5 demonstrates high recall, generates fluent corrections, and generally preserves sentence semantics.  However, {\em human} evaluation for both English and Russian reveals that, despite its strong error-detection capabilities, GPT-3.5 struggles with several error types, including punctuation mistakes, tense errors, syntactic dependencies between words, and lexical compatibility at the sentence level.
 \\ \newline  }

\begin{document}

\maketitleabstract

\section{Introduction}

Recent advances in large language models (LLM)---e.g., GPT-3~\cite{brown2020language}, GPT-4~\cite{openai-report}, LLaMA~\cite{touvron2023llama},  Gemini~\cite{team2023gemini}, Mixtral~\cite{jiang2024mixtral}, etc.---have resulted in strong performance on various NLP tasks in zero-shot or few-shot settings.
%These promising results naturally motivate exploration of LLM's performance on the task of Grammatical Error Correction.
Recent work~\cite{coyne2023analysis,coyneanalyzing,fang2023chatgpt,loem-etal-2023-exploring} has assessed several GPT-3.5 models on the task of grammatical error correction (GEC), mostly for English, and compared with other state-of-the-art models, showing great potential of GPT-3.5 for GEC. %We extend previous experiments on GEC using GPT models, including ChatGPT-3.5~\footnote{https://openai.com/blog/chatgpt} and  GPT-3.5~\cite{} which are designed to follow instructions and provide detailed responses.

We experiment with multiple languages in several settings: zero-shot setting, including zero-shot chain-of-thought (CoT)~\cite{kojima2022large}, fine-tuning GPT-3.5 models, and using GPT-3.5\footnote{https://openai.com/blog/chatgpt} to re-rank hypotheses generated by other GEC models.
To our knowledge, GPT-3.5 has not been assessed for GEC in all three of these settings previously.  Our experiments were performed at the sentence level using OpenAI's GPT-3.5-turbo\footnote{Version \texttt{gpt-3.5-turbo-0613}} models through the official OpenAI API.\footnote{https://platform.openai.com/docs/api-reference}

% Check below
%using reference-based metrics.\comment{??? is THIS the challenge -- in the following sentence ?  If so, then maybe better end this sentence with : and continue to the challenge }
Since GPT-3.5 is known to suggest fluent corrections with high recall~\cite{fang2023chatgpt,wu2023chatgpt}, reference-based metrics tend to under-estimate the model's true performance and provide only a partial estimation of it, as detailed in~\cite{rozovskaya-roth-2021-good}.  However, we lack other metrics suitable for verifying that GPT-3.5 indeed corrects grammatical errors and does not rewrite the input into a coherent yet irrelevant sentence.  These increasing challenges in evaluating the performance of GEC models have motivated us to use manual and reference-free evaluation, as well as measuring semantic similarity between the input sentences and the corrected sentences.

We address the following research questions: (RQ1)~Can GPT-3.5 be used consistently for GEC across different languages, in different settings, such as zero-shot and fine-tuning?
(RQ2)~Does GPT-3.5 alter the semantics of the source sentence while attempting to correct it? Based on our experiments, we make the following observations:

(A) In a zero-shot setting, GPT-3.5 achieves higher recall than other GEC models for the majority of the tested languages (except Czech and Ukrainian).  However, as manual evaluation of sentences corrected by GPT-3.5 for English and Russian shows, corrected sentences can deviate significantly from the input.  Fine-tuning performance is much lower for all tested languages, with numerous sentences remaining uncorrected.

(B) Using GPT-3.5 to re-rank correction hypotheses generated by other GEC models results in an increased recall, as the model ranks more fluent corrections higher.  Re-ranking hypotheses with GPT-3.5 beats GPT-3.5's performance in the zero-shot setting for GEC across all evaluated languages.  This indicates that GPT-3.5 may be more effectively used as a re-ranker in scenarios where other smaller fine-tuned GEC models are available.

(C) A detailed manual evaluation of the corrections suggested by GPT-3.5 showed that the model is able to correct errors and to handle long sequences and reversed word order.  However, it often struggles with correcting punctuation, article, preposition, and verb tense errors in English; it struggles with correcting errors in punctuation, government relations between words, and lexical compatibility in Russian.

(D) Reference-free evaluation of GPT-3.5 performance in a zero-shot setting---using pre-trained language models (LMs), Scribendi scores~\cite{islam-magnani-2021-end}, and evaluation of semantic similarity between sentence embeddings---reveal that GPT-3.5 corrections tend to deviate from the source sentences most in Czech, German, Spanish, and Ukrainian.  The model modifies the input for these languages more liberally, which does not necessarily improve performance.  For Russian, the results indicate that GPT-3.5 can correct more errors than any other GEC model, although its suggestions often semantically diverge from the source sentences.  The performance for English appears more promising since the suggested corrections remain semantically similar to the input sentences.  Additionally, over 50\% of the sentences corrected by GPT-3.5 are evaluated by LMs as more fluent than the reference sentences.  For Swedish, GPT-3.5 exhibits relatively high precision compared to other languages, with its correction hypotheses staying semantically close to the input, which may, though, result from under-corrected errors.

% (F) Prompts significantly influence performance in machine translation (MT)~\cite{gu2023linguistically,jiao2023chatgpt}, GEC~\cite{fang2023chatgpt}, and other NLP tasks~\cite{lai2023chatgpt}.  However, we find that the effectiveness of specific prompt strategies may not consistently extend across different languages, datasets, or even when applied to the same dataset.  For instance, we were unable to reproduce the performance achieved by using the CoT prompting technique for GEC on the CoNLL-2014 and BEA-19 test sets, as described by~\cite{fang2023chatgpt}.

The paper is structured as follows.  Section~\ref{related} provides a brief overview of related work on GEC.  Section~\ref{experiments} details the datasets and describes experiments conducted in three settings: zero-shot, fine-tuning, and re-ranking of correction hypotheses.  Section~\ref{evaluation} discusses the results of these experiments.  Section~\ref{analyses} focuses only on the corrections generated by GPT-3.5 in the zero-shot setting.  These corrections are evaluated manually, with LMs, and with the Scribendi Score.  This section also examines the semantic similarity between the corrections generated by GPT-3.5 and the original sentences.  Finally, Section~\ref{conclusion} summarizes the findings and outlines directions for future research.

\section{Related Work}
\label{related}

Grammatical error correction is usually approached as a monolingual sequence-to-sequence task: rewrite an incorrect sentence into its corrected version.  This allows us to use models developed for machine translation~\cite{ji-etal-2017-nested,junczys-dowmunt-etal-2018-approaching,chollampatt2018mlconv,yuan-etal-2019-neural,grundkiewicz-etal-2019-neural,zhao-etal-2019-improving,kiyono-etal-2019-empirical,choe-etal-2019-neural,kaneko-etal-2020-encoder, rothe-etal-2021-simple}. As the method requires a large amount of data, models are commonly pre-trained on synthetic datasets. There are various techniques for augmenting error-free data, e.g., by adding random word perturbations, Wikipedia edits, inserting errors by back translation, or imitating real error patterns~\cite{kiyono-etal-2019-empirical,grundkiewicz-etal-2019-neural,choe-etal-2019-neural,takahashi-etal-2020-grammatical,li-he-2021-data}. Another approach involves tagging sentences with token transformation labels and subsequently editing them based on the predicted transformations~\cite{malmi-etal-2019-encode,stahlberg-kumar-2020-seq2edits,omelianchuk-etal-2020-gector,tarnavskyi-etal-2022-ensembling}. More work on the GEC is reviewed in~\cite{bryant2022grammatical}.

GEC can be approached using Generative Adversarial Networks---GANs~\cite{raheja-alikaniotis-2020-adversarial}, where the neural MT-based generator fixes errors, while the discriminator judges the quality of corrections. ~\citet{parnow-etal-2021-grammatical} presented a GAN-like architecture that consists of an error sequence labeler as a generator and an error detector as a discriminator. %These works approach the problem of scarce training data differently from the classical way of generating synthetic datasets by introducing errors into error-free corpora. %For example, in~\cite{parnow-etal-2021-grammatical}, synthetic erroneous sentences will increasingly represent real error data with more training.

\citet{korotkova2019grammatical} investigated zero-shot GEC using Transformer-based MT models. \citet{fang2023chatgpt} assessed ChatGPT's performance for Chinese, English, and German GEC on the sentence level, and for English on the document level. %The authors performed a human evaluation of the model's output on different criteria.
ChatGPT showed the highest scores on fluency and ability to generate correct results beyond what is indicated by the reference in comparison with other GEC models, such as GECToR~\cite{omelianchuk-etal-2020-gector}, T5-based models~\cite{rothe-etal-2021-simple}, and the commercial system Grammarly. \citet{coyne2023analysis} tested GPT-3.5 and GPT-4 performance on English GEC across zero-shot and few-shot settings and showed the tendency of the models to over-edit, changing the meaning of the sentence during correction.

\section{Experiments}
\label{experiments}

\subsection{Data}

We use the official datasets for all languages.  For English, we use ConLL-14~\cite{ng-etal-2014-conll}, BEA-2019~\cite{bryant-etal-2019-bea}, and JGLEG~\cite{napoles-etal-2017-jfleg} test sets only for evaluation.  The datasets FCE~\cite{yannakoudakis-etal-2011-new} and W\&I~\cite{bryant-etal-2019-bea} were combined for fine-tuning.
We use the test sets from AKCES-GEC~\cite{naplava-straka-2019-grammatical} for Czech, COWS-L2H~\cite{yamada2020cows} for Spanish, and UA-GEC~\cite{syvokon-etal-2023-ua} for Ukrainian for evaluation.
We use Falco-MERLIN~\cite{boyd-etal-2014-merlin} and RULEC-GEC~\cite{rozovskaya-roth-2019-grammar} for German and Russian, respectively: the training sets were used for fine-tuning, and test sets for evaluation.
For Swedish, we use SweLL-gold corpus~\cite{volodina2019swell}, which contains 502 essays written by adult learners of Swedish at various proficiency levels.  The essays are pseudo-anonymized and manually annotated.  Given that this corpus includes some personal information, it is under GDPR protection.\footnote{https://gdpr-info.eu/}
For evaluation, we shuffle the sentences in the entire corpus and randomly sample 60\% of them.  Statistics on the datasets are provided in Table~\ref{table:datasets}.

\begin{table}[t!]
  \begin{center}
  \scalebox{0.93}{
      \begin{tabular}{c|l|r|r|r}
\hline
         & \textbf{Dataset} & \textbf{Train} & \textbf{Dev}  & \textbf{Test} \\
\hline
        CZ & AKCES-GEC & 42 210 & 2 485 & 2 676 \\
        DE & Falko-MERLIN & 19 237 & 2 503 & 2 337 \\
        EN & BEA-19 & - & - & 4 477 \\
        EN &  CoNLL-14 & - & - & 1 312 \\
        EN &  JGLEG & - & - & 747 \\
        EN & FCE, W\&I & 59 941 & - & - \\
        ES & COWS-L2H & 6 159 & 1 230 & 822 \\
        RU & RULEC-GEC & 4 980 & 2 500 & 5 000 \\
        SV & SweLL-gold$^*$ & - & - & 8 553  \\
        UA & UA-GEC & 31 046 & - & 2 704 \\
\hline
      \end{tabular}
      }
    \caption{Datasets used for fine-tuning and evaluation. We sampled 60\% of the sentences from shuffled SweLL-gold.}
    \label{table:datasets}
  \end{center}
\end{table}

\subsection{Settings}
\label{settings}

\paragraph{Zero-shot GEC:} For each input sentence, we prompted the model to follow the minimum-edit principle for correcting erroneous sentences.  GPT-3.5 tends to provide feedback on the input's correctness or to comment on the corrections.  Therefore, we followed the suggestion of~\citet{fang2023chatgpt} and added input and output tags to format the model's output.

We experimented with two types of prompts. The first type is GEC instruction:

Prompt 1: ``{\tt Provide a grammatical correction for the following sentence indicated by <input> ERROR </input> tag, making only necessary changes.  If the input text is already correct, return it unchanged.  Output the corrected version directly without any comments and explanations. Remember to format your corrected output with the tag <output> Your Corrected Version </output>. Please start: <input> ERROR </input>}''

The second prompt is a chain-of-thought~\cite{kojima2022large} prompt suggested by~\citet{fang2023chatgpt}.  It requires the model to follow particular steps of input comprehension:

Prompt 2: ``{\tt Please identify and correct any grammatical errors in the following sentence indicated by <input> ERROR </input> tag, you need to comprehend the sentence as a whole before identifying and correcting any errors step by step while keeping the original sentence structure unchanged as much
as possible. Afterward, output the corrected version directly without any
explanations. Remember to format your corrected output results with the
tag <output> Your Corrected Version </output>. Please start: <input> ERROR
</input>}''

In both cases, we provide the model with information about its role:

``{\tt You are a LANGUAGE\footnote{Variable for the target language.} grammatical error correction tool that can identify and correct grammatical errors in a text.}''

We also require the model to {\em not} translate input into English in case it is in other languages: ``{\tt Do not translate input sentences into English.}''

For all languages, we utilize English prompts for consistency. Our objective is to standardize prompts to reduce variation in the input, focusing solely on the target language, thereby ensuring more controlled outcomes.

\paragraph{Fine-tuning GPT-3.5:}

Fine-tuning was performed for English, German, and Russian. We use the GPT-3.5-turbo model for all three languages, without altering any parameters.\footnote{https://platform.openai.com/docs/guides/fine-tuning}  Training examples were presented in a conversational chat format, with the same system role as for the zero-shot setting.  We experiment with using (1) a short prompt message of ``{\tt Improve grammar in the following sentence: ERROR}''\footnote{For Russian and German: ``{\tt Improve grammar in the following sentence, do not translate into English: ERROR}''. \texttt{ERROR} is a variable to replace with an input sentence.} and (2) a longer message, which is the same as Prompt 2 for the zero-shot setting.

\begin{table*}
  \begin{center}
  \scalebox{0.90}{
  \begin{tabular}{l|l|l|cccc}
\hline
     \textbf{Lang.} & \textbf{Dataset} & \textbf{Model} & $P$ & $R$ &  $F_{0.5}$ & $\sigma$\\
\hline
\hline
     CZ & AKCES-GEC & \underline{ChatGPT (zero-shot)} & 68.5 & 64.1 & 67.6 & 1.9 \\
     & & Transformer~\cite{naplava-straka-2019-grammatical} & \textbf{84.2} & \textbf{66.7} & \textbf{80.0} & - \\
     \hline
    DE & Falco-MERLIN & ChatGPT (zero-shot)~\cite{fang2023chatgpt} & 59.9 & \textbf{63.9} & 60.7 &-  \\
    & &  \underline{ChatGPT (zero-shot)} & 63.5 & 63.2 & 63.4 & 0.7 \\
    % & & mT5~\cite{fang2023chatgpt} & 75.4 & 55.1 & 70.2 \\
    & & Transformer~\cite{naplava-straka-2019-grammatical} & \textbf{78.2} & 59.9 & 73.7 & - \\
    & & T5 xxl~\cite{rothe-etal-2021-simple} & - & - & \textbf{76.0} & - \\
\hline
    EN & BEA-19 (test) & ChatGPT (zero-shot)~\cite{fang2023chatgpt} & 32.1 & \textbf{70.5} & 36.1 & - \\
    & &  \underline{ChatGPT (zero-shot)} & 47.4 & 64.6 & 50.0 & 3.5 \\
    & &  \underline{ChatGPT (rerank)} & 66.2 & 61.5 & 65.2 & - \\
    % & & GECToR~\cite{omelianchuk-etal-2020-gector} & 76.7 & 57.8 & 71.9 \\
    % & & T5 large~\cite{fang2023chatgpt} & 73.4 & 67.0 & 72.0 \\
    % & & T5 xxl~\cite{rothe-etal-2021-simple} & - & - & 75.9 \\
    & & ESC~\cite{qorib-etal-2022-frustratingly} & \textbf{86.7} & 60.9 & \textbf{79.9} & - \\
    &  CoNLL-14 & ChatGPT (zero-shot)~\cite{fang2023chatgpt} & 50.2 & \textbf{59.0} & 51.7 & - \\
    & & \underline{ChatGPT (zero-shot)} & 55.8 & 58.5 & 56.3 & 0.7 \\
    & & \underline{ChatGPT (rerank)} & 59.9 & 52.4 & 58.3 & - \\

    % & & GECToR~\cite{omelianchuk-etal-2020-gector} & 75.6 & 44.5 & 66.3 \\
    % & & T5 large~\cite{fang2023chatgpt} & 72.2 & 51.4 & 66.8 \\
    % & & T5 xxl~\cite{rothe-etal-2021-simple} & - & - & 68.9 \\
    & & ESC~\cite{qorib-etal-2022-frustratingly} & \textbf{81.5} & 43.8 & \textbf{69.5} & - \\
    \hline
    ES & COWS-L2H & mT5 & \textbf{29.0} & 10.8 & 21.7 & - \\
    & & \underline{ChatGPT (zero-shot)} &  28.1 & \textbf{40.8} & \textbf{29.9} & 2.2 \\
    \addlinespace[2pt]
    \hline
    RU & RULEC-GEC &  \underline{ChatGPT (zero-shot)} & 29.2 & \textbf{51.5} & 32.0 & 3.0 \\
    & & \underline{ChatGPT (rerank)}  & 39.5 & 43.7 & 40.3 & - \\
    & & T5 xxl~\cite{rothe-etal-2021-simple} & - & - & 51.6 & - \\
    & & ruGPT large + re-ranking~\cite{sorokin-2022-improved} & \textbf{73.3} & 27.3 & 55.0 & - \\
    & & RuT5 large + re-ranking~\cite{katinskaia-yangarber-2023-grammatical} & - & - & \textbf{68.2} & - \\
\hline
    SV & SweLL-gold$^*$ &  \underline{ChatGPT (zero-shot)} & \textbf{60.1} & \textbf{49.4} & \textbf{57.6} & 1.8 \\
\addlinespace[2pt]
\hline
    UA & UA-GEC &  \underline{ChatGPT (zero-shot)} & 25.8 & 36.2 & 27.4 & 0.6 \\
     & & mT5 large~\cite{syvokon2023unlp} & \textbf{76.8} & \textbf{61.4} & \textbf{73.1} & - \\
\hline
  \end{tabular}
}
  \caption{Performance measured with ERRANT for the BEA-19 dataset, and with the M2 Scorer for other datasets.
  Best scores across a language (across datasets for English) are in {\bf bold}.
  State-of-the-art models are provided for comparison.
  Our results are \underline{underlined}.  All \texttt{ChatGPT} models in the table are GPT-3.5 versions. $\sigma$ denotes standard deviation for $F_{0.5}$ measure across 3 requests.}
  \label{table:zero}
  \end{center}
\end{table*}

\paragraph{Re-ranking with GPT-3.5:}
\label{re-ranking}
For re-ranking, we use pre-existing GEC models: a publicly available encoder-decoder model\footnote{https://github.com/butsugiri/gec-pseudodata} trained on synthetic data for English~\cite{kiyono-etal-2019-empirical}; T5-large model,\footnote{https://huggingface.co/ai-forever/ruT5-large} which we pre-train on synthetic data and then fine-tune on the RULEC-GEC training set for Russian, following~\citet{rothe-etal-2021-simple}.
Synthetic data was generated from WMT News Crawl monolingual training data~\cite{bojar-etal-2017-findings} using Aspell confusion sets, following the approach of~\citet{grundkiewicz-etal-2019-neural}.
 We generated 10M sentences, with the same parameters as in~\cite{naplava-straka-2019-grammatical}.  The T5 model was pre-trained until convergence with the following parameters: 3 GPU V100, pre-training for 1.48M steps with batch size 6, weight decay 0, learning rate 5e-5.

We generate 5 hypotheses for each source sentence using beam search decoding.
The list of hypotheses for each sentence was provided to GPT-3.5 together with a prompt.

Prompt 3: ``{\tt Re-rank the following sentences according to their grammatical correctness: {\tt LIST}.\footnote{A variable to replace with a list of hypotheses.} Do not explain your choice. Numerate re-ranked sentences following the format:}

{\tt 1. Sentence A}

{\tt 2. Sentence B}

{\tt etc.}

{\tt Do not provide any comments, notes, or translations. Do not change the input sentences.}''

\section{Evaluation}
\label{evaluation}

\paragraph{Zero-shot:} Given the variability of GPT-3.5's output for the same input on each request, we averaged the performance over three requests per input sentence, see {\tt ChatGPT (zero-shot)} in Table~\ref{table:zero}.  Prompt 2 showed better performance for all languages than Prompt 1.  Thus, Table~\ref{table:zero} shows performance with the Prompt 2.
All $F_{0.5}$ scores for GPT-3.5 are higher than those of~\citet{fang2023chatgpt} in zero-shot CoT settings, though their recall is higher for English and German.  This difference is likely due to the fact that the model may have evolved in the meantime.  For Spanish, we compared the performance of GPT-3.5 with the results presented in this \href{https://diligent-raver-536.notion.site/Spanish-Grammatical-Error-Correction-88d0f0d1d090412baf4c52cdf87a0468}{GEC report}.\footnote{https://github.com/abhisaary/spanish\_gec}
We are not aware of any Swedish GEC model to compare with, whose performance was measured by ERRANT or the M2 Scorer.

\begin{table}[t!]
  \begin{center}
  \scalebox{0.94}{
  \begin{tabular}{c|ccc}
\hline
      \textbf{Annotator} & $P$ & $R$ &  $F_{0.5}$ \\
\hline
    A & 67.2 &	58.8 &	65.3 \\
    B & 65.3 & 	59.2 &	64.0 \\
    C & 69.6 &	68.5 &	69.4 \\
    D & 61.9 &	70.0 &	63.4 \\
\hline
  \end{tabular}}
  \caption{Performance of GPT-3.5 in zero-shot setting measured with ERRANT for BEA-19 test set, reported separately for all 4 reference annotations.}
  \label{table:annotators}
  \end{center}
\end{table}

\begin{table}
  \begin{center}
  \scalebox{0.94}{
  \begin{tabular}{c|l|ccc}
\hline
      \textbf{Lang.} & \textbf{Dataset}  & $P$ & $R$ &  $F_{0.5}$ \\
\hline
    EN & CoNLL-14 & 45.6 & 41.4 & 44.7 \\
     & BEA-19 (test) & 40.4 & 51.5 & 42.2 \\
    DE & Falco-MERLIN & 54.5 & 42.9 & 52.5 \\
    RU & RULEC-GEC & 43.5 & 17.2 & 33.3 \\
\hline
  \end{tabular}
  }
  \caption{Results of fine-tuning GPT-3.5 for GEC. For BEA-19, the performance was measured using ERRANT, and with M2 Scorer for other datasets.}
  \label{table:finetune}
  \end{center}
\end{table}

For all languages, except Czech and Ukrainian, GPT-3.5 has higher recall than all other models, but lower precision.  High recall indicates strong error-correction capabilities, but the proposed corrections may differ from the reference corrections used for automatic evaluation.
For example, as shown in Table~\ref{table:annotators}, GPT-3.5 performance on English varies when we run evaluations against the gold-standard references within the BEA-19 test set provided by different annotators.
Such variability in performance highlights the limitations of reference-based evaluation: to obtain an accurate assessment of GEC performance, it is essential to have {\em multiple} gold-standard references for each input sentence across all examined languages.

As shown in Table~\ref{table:zero}, none of the GPT-3.5 models achieved the highest $F$-score across all languages, except for Spanish.  This finding aligns with the results of~\citet{zhang2023multi}, who evaluated the performance of fine-tuned LLaMA on various writing-related tasks, including assessment of grammaticality.  The authors demonstrated that a fully fine-tuned LLaMA model with 13B parameters achieves a comparable performance to that of lightweight RoBERTa and T5 models.  The latter models are much faster and cheaper both in the training and the inference phases.

\vspace{4pt}
{\bf Fine-tuning:} Among the two tested prompts, the short prompt achieved higher performance, see results in Table~\ref{table:finetune}.  The performance of the fine-tuned models is significantly lower than other settings for all tested languages.
Manual inspection of the suggested corrections revealed that numerous sentences in the test remain unchanged.
Given that fine-tuning aims to instruct the model on the expected corrections (i.e., making minimal changes without altering the sentence's meaning), we hypothesize that fine-tuning for GEC requires more careful preparation of training data and prompt selection.
Notably, the results for Russian exhibit very low recall, which could be due to the relatively small size of the training set and the presence of some number of uncorrected errors in the data.

\vspace{4pt}
{\bf Re-ranking:} Re-ranking performance depends on the performance of the GEC models that are used to generate the hypotheses.  In Table~\ref{table:reranking}, we report the performance of English and Russian GEC models, mentioned in Subsection~\ref{settings}.  Performance before re-ranking is marked with \xmark; performance after re-ranking---with \cmark.
As we can see, re-ranking with GPT-3.5 increases only recall, while decreasing precision.
A possible explanation for this is that the model ranks the more fluent hypotheses with fewer errors higher, even if they deviate from the source sentences.

\begin{figure}
\begin{center}
  \includegraphics[scale=0.45]{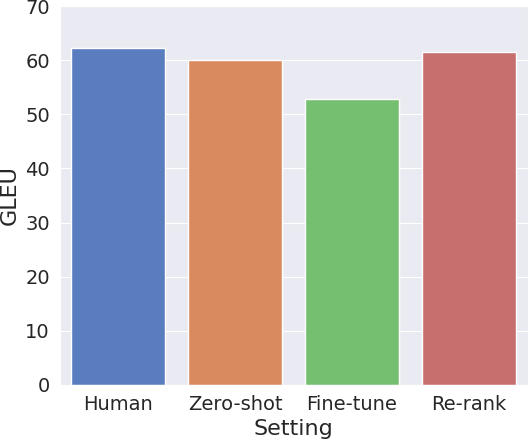}
  \caption{Performance for JFLEG, measured in GLUE.}
  \label{fig:gleu}
\end{center}
\end{figure}

Table~\ref{table:zero} shows that re-ranking correction hypotheses with GPT-3.5 performs better than in the zero-shot setting for GEC for English and Russian.  Evaluation on JFLEG, a corpus for evaluating GEC systems with respect to fluency and grammatically, shows very high performance of GPT-3.5 as a re-ranker for English in comparison with a human-level evaluation score, see Figure~\ref{fig:gleu}.  These observations indicate that GPT-3.5 may be more effectively used as a re-ranker in scenarios where other smaller fine-tuned GEC models are available and over-corrections/hallucinations are not tolerated.

\begin{table}[t!]
  \begin{center}
  \scalebox{0.85}{
  \begin{tabular}{c|l|c|ccc}
\hline
      \textbf{Lang.} & \textbf{Dataset} & \textbf{Re-rank} & $P$ & $R$ &  $F_{0.5}$ \\
\hline
    EN & BEA-19 (test) & \xmark & 66.2 & 61.5	& 65.2 \\
    & BEA-19 (test) & \cmark & 47.4 & 64.6 & 50.0 \\
    % & CONLL-2014 (no alt) & 56.86 & 50.34 & 55.43 \\
    & CoNLL-14 & \xmark & 68.6 & 44.9 & 62.1 \\
    & CoNLL-14 & \cmark & 59.9 & 52.4 & 58.3 \\
    % & FCE & 46.20 & 47.33 & 46.42 \\
    % & JFLEG & & & \\ % TODO: run reranking for generated hyps
    RU & RULEC-GEC & \xmark & 62.5 & 27.2 & 49.6 \\
    & RULEC-GEC & \cmark & 38.4 & 42.5 & 39.1 \\
\hline
  \end{tabular}
  }
  \caption{Re-ranking hypotheses with GPT-3.5.
  For BEA-19, performance was measured with ERRANT, and with M2 Scorer for other datasets.}
  \label{table:reranking}
  \end{center}
\end{table}

\begin{table*}[ht!]
\begin{center}
\scalebox{0.81}{
\begin{tabular}{l}
\hline
\textbf{Original:}\hspace{4mm} When I have spare time, I often gather my friends to watch basketball match on television. \\
\textbf{Corrected:}\hspace{0.7mm} When I have spare time, I often gather my friends to watch {\color{BrickRed} basketball match} on television. \\
\textbf{Reference:}\hspace{0.21ex} When I have spare time, I often gather together my friends to watch {\color{Blue} basketball matches} on television. \\
\midrule
% \textbf{Original:}\hspace{3mm} When I arrived my uncle was waiting for me and I gave him a big hug. \\
% \textbf{Corrected:} When I arrived my uncle was waiting for me and I gave him a big hug. \\
% \textbf{Reference:}\hspace{0.21ex} When I arrived{\color{Blue},} my uncle was waiting for me and I gave him a big hug. \\
% \midrule
\textbf{Original:}\hspace{4mm} He is clearly aware of the dangers and brutalism of the sport, which is possibly why he enjoys it so much. \\
\textbf{Corrected:}\hspace{0.7mm} He is clearly aware of the dangers and {\color{BrickRed} brutalism} of the sport, which is possibly why he enjoys it so much. \\
\textbf{Reference:}\hspace{0.21ex} He is clearly aware of the dangers and {\color{Blue} brutality} of the sport, which is possibly why he enjoys it so much. \\
\midrule
\textbf{Original:}\hspace{4mm} However, things do n't happened like what I hoped. \\
\textbf{Corrected:}\hspace{0.7mm} However, things {\color{BrickRed} don't happen like} what I hoped. \\
\textbf{Reference:}\hspace{0.21ex} However, things {\color{Blue} didn't happen like} I hoped. \\
\midrule
\textbf{Original:}\hspace{4mm} I can speak English and I am a lovely, energetic and hardworking person besides I have recommended paper...  \\
\textbf{Corrected:}\hspace{0.7mm} I can speak English, and I am a lovely and hardworking person. Besides, I have {\color{BrickRed} a recommended paper}... \\
\textbf{Reference:}\hspace{0.21ex} I can speak English and I am a lovely and hardworking person. In addition, I have {\color{Blue} a reference}... \\
\midrule
\textbf{Original:}\hspace{4mm} \textcyr{которые образованы о обеспечивании государства высокого места в международной иерархии} \\
\textbf{Corrected:}\hspace{0.7mm} \textcyr{которые {\color{BrickRed} разбираются в обеспечении государства высокого места} в междунарожной иерархии} \\
\textbf{Reference:}\hspace{0.21ex} \textcyr{которые {\color{Blue} образованны в области обеспечения государству высокого места} в международной иерархии} \\
\hspace{10ex} ({\em who are really educated in the field of providing the state with a high place in the international hierarchy}) \\
\midrule
\textbf{Original:}\hspace{4mm} \textcyr{Она всегда в модной одежде и ведет себя властью.} \\
\textbf{Corrected:}\hspace{0.7mm} \textcyr{Она всегда в модной одежде и {\color{BrickRed} ведет себя с властью}.} \\
\textbf{Reference:}\hspace{0.21ex} \textcyr{Она всегда в модной одежде и {\color{Blue} ведет себя властно}}. \\
\hspace{10ex} (\textit{She is always in fashionable clothes and acts bossy.}) \\
\midrule
\textbf{Original:}\hspace{4mm} \textcyr{Я хочу привести яркий пример, о том, что может случиться...} \\
\textbf{Corrected:}\hspace{0.7mm} \textcyr{Я хочу привести яркий {\color{BrickRed} пример о том}, что может случиться...} \\
\textbf{Reference:}\hspace{0.21ex} \textcyr{Я хочу привести яркий {\color{Blue} пример того}, что может случиться...} \\
\hspace{10ex} (\textit{I want to give a vivid example of what could happen...}) \\
% \midrule
% \textbf{Corrected:} \textcyr{...мы призываем вас подумать о том, что {\color{BrickRed} водные дела} также важны в Орегоне} \\
% \textbf{Reference:}\hspace{0.21ex} \textcyr{...мы призываем вас подумать о том, что {\color{ForestGreen} водные проблемы} также важны в Орегоне} \\
% \hspace{9ex} (\textit{...we encourage you to consider that water issues are also important in Oregon}) \\
% \midrule
% \textbf{Original:}\hspace{3mm} \textcyr{В результате, большая часть мир жаждет, или пьет нечистую воду и болеет.} \\
% \textbf{Corrected:} \textcyr{В результате {\color{BrickRed} большой части мира люди жаждут} или пьют нечистую воду и болеют.} \\
% \textbf{Reference:}\hspace{0.21ex} \textcyr{В результате{\color{Blue}, большая часть мира испытывает недостаток питьевой воды} или пьет нечистую воду и болеет.} \\
% \hspace{9ex}(\textit{As a result, much of the world lacks drinking water, or drinks unclean water and gets sick.}) \\
\midrule
\textbf{Original:}\hspace{4mm} \textcyr{В курсе Основы светской этики учатся долг и совесть, принципы морали..} \\
\textbf{Corrected:}\hspace{0.7mm} \textcyr{В курсе ``Основы светской этики'' {\color{BrickRed} учатся долгу и совести, принципы морали}..} \\
\textbf{Reference:}\hspace{0.21ex} \textcyr{В курсе ``Основы светской этики'' {\color{Blue} изучаются долг и совесть, принципы морали}...} \\
\hspace{10ex} (\textit{The course ``Fundamentals of secular ethics'' covers duty and conscience, the principles of morality...}) \\
\hline
\end{tabular}}
\caption{Examples of erroneous corrections (see \textbf{Corrected}) which were suggested by GPT-3.5 in a zero-shot setting for English and Russian. Red denotes errors, blue denotes corrections suggested by the annotators.}
\label{table:notcor}
\end{center}
\end{table*}

\section{Analysis of Zero-shot Output}
\label{analyses}

\paragraph{Zero-shot (Manual evaluation):} We carry out a preliminary manual evaluation for English and Russian.  We randomly pick 200 sentences for each language: 70\% of the sentences corrected by GPT-3.5, and 30\% reference sentences.  Two annotators, native speakers of both languages, were not informed about the origin of the evaluated sentences.  Their task was to assess whether a sentence is grammatically acceptable.

The evaluation shows that 86.5\% of the English sentences and 80\% of the Russian sentences are acceptable for the annotators.  For English, the majority of the errors are related to usage of hyphen, preposition, article, and tense.  For Russian, most errors relate to punctuation, tense, and lexical compatibility.  The model struggles to correct errors in constructions with government relations: errors in the usage of case of nouns and adjectives governed by verbs, or in prepositions required by particular verbs.

Table~\ref{table:notcor} shows examples of erroneous corrections proposed by GPT-3.5 and references suggested by the annotators.  Interestingly, although GPT-3.5 tends to propose fluent corrections, the examples demonstrate an absence of agreement between proposed corrections and the semantics of the whole sentence.

Results of manual evaluation agree with error-type performance evaluated using ERRANT, see detailed error performance by error types for English (Table~\ref{table:types1}) and Russian (Table~\ref{table:types2}).  Red denotes error types that were noted as problematic by the annotators.

\vspace{4pt}
{\bf Evaluation with LM:}
For automatic reference-free evaluation, we explore the use of pre-trained LMs. The use of pre-trained LMs for assessing grammaticality has been proposed by several researchers~\cite{linzen-etal-2016-assessing,goldberg2019assessing}. Prior work has described incorporating LMs as components into GEC systems to score corrections, or to filter out low-quality examples from data~\cite{lichtarge-etal-2020-data,yasunaga-etal-2021-lm}. The core approach described in~\cite{yasunaga-etal-2021-lm} involves using an LM critic to obtain realistic pairs of grammatical and erroneous sentences from unlabeled data. The role of the critic is to assess whether an input sentence is correct by utilizing the LM's probability score.  We adopt \newcite{yasunaga-etal-2021-lm}'s intuition, that for a grammatical sentence, $x_{good}$, and its ungrammatical version with the same
intended meaning, $x_{bad}$, we expect $p(x_{bad}) < p(x_{good})$, where $p$ denotes the LM's probability normalized by the sentence length.

\begin{table}[ht!]
\begin{center}
\scalebox{0.85}{
  \begin{tabular}{l|ccc}
\hline
      \textbf{Error type} & $P$ & $R$ &  $F_{0.5}$ \\
\hline
ADJ	& 48.8 & 46.5 &	48.3 \\
ADJ:FORM &	71.4 &	76.9 &	72.5  \\
ADV	& 43.9 & 51.3 &	45.3  \\
CONJ	& 43.6 &	53.1 &	45.2  \\
CONTR &	92.9	& 86.7 &	91.6  \\
DET	& 69.1 & 70.9 &	69.5  \\
MORPH &	75.7 &	72.8 &	75.1  \\
 {\color{BrickRed} NOUN } & {\color{BrickRed} 36.6} &  {\color{BrickRed}  37.5 } &  {\color{BrickRed} 36.8  }\\
NOUN:INFL & 84.2 &	84.2 & 84.2  \\
NOUN:NUM &	68.7 &	83.2 & 71.2  \\
NOUN:POSS	& 78.4	& 52.7 & 71.4  \\
ORTH &	86.3 & 62.6 & 80.2  \\
OTHER &	43.9 & 31.9 & 40.8  \\
PART & 68.6 & 82.7 & 71.0  \\
 {\color{BrickRed} PREP } &	 {\color{BrickRed} 64.6 } &  {\color{BrickRed}  67.2 } &  {\color{BrickRed} 65.2 }  \\
PRON & 65.0 & 59.8 &	63.9  \\
PUNCT &	72.3 & 59.6 & 69.3  \\
SPELL & 79.1 &	78.2 & 78.9  \\
 {\color{BrickRed} VERB } &  {\color{BrickRed} 	53.0 } &  {\color{BrickRed}  42.3 } &  {\color{BrickRed}  50.5 }  \\
VERB:FORM &	73.7 & 81.6 & 75.1  \\
VERB:INFL & 100.0 & 87.5 & 97.2 \\
VERB:SVA &	80.0 &	90.0 & 81.8  \\
 {\color{BrickRed} VERB:TENSE} &  {\color{BrickRed} 59.4 } &  {\color{BrickRed} 58.8 } &  {\color{BrickRed} 59.3 }  \\
 {\color{BrickRed} WO } &  {\color{BrickRed}  65.2 } &  {\color{BrickRed} 48.9 } &  {\color{BrickRed} 61.1 } \\
\hline
  \end{tabular}}
  \caption{Error-type performance of GPT-3.5 in zero-shot setting for BEA-19 test set (English), measured using ERRANT.}
  \label{table:types1}
  \end{center}
\end{table}

\begin{table}[ht!]
  \begin{center}
  \scalebox{0.85}{
  \begin{tabular}{l|rrr}
\hline
      \textbf{Error type} & $P$ & $R$ &  $F_{0.5}$ \\
\hline
 {\color{BrickRed} ADJ }            &  {\color{BrickRed}     8.5}  &  {\color{BrickRed}  23.4 } &  {\color{BrickRed}   9.7 } \\
 {\color{BrickRed} ADJ:CASE }        &   {\color{BrickRed}    20.2 } &  {\color{BrickRed}   59.1  } &  {\color{BrickRed}   23.3 } \\
ADJ:COMP\_FORM    &     20.0  &    25.0  &   20.8 \\
ADJ:FULL/SHORT   &     32.3 &  56.6  &  35.3 \\
ADJ:GEN        &     40.8 &  62.5 &  43.9 \\
ADJ:INFL  &  14.3  & 60.0  &  16.9 \\
ADJ:NUM    &        30.0  &   58.1 &  33.2 \\
ADV      &          11.8  & 25.4 &  13.3 \\
CCONJ    &      6.8  & 30.6 &  7.9 \\
MORPH    &          43.1 &  63.5  & 46.1 \\
 {\color{BrickRed} NOUN }       &   {\color{BrickRed}  16.5 } &  {\color{BrickRed} 28.2 } &   {\color{BrickRed} 17.9 } \\
 {\color{BrickRed}  NOUN:CASE }  &  {\color{BrickRed} 37.2 } &  {\color{BrickRed} 69.5 } &  {\color{BrickRed} 41.0 } \\
NOUN:INFL   &        66.7 &  80.0  &   68.9 \\
NOUN:NUM     &       28.7 &  53.9 &  31.6 \\
NUMERAL           &    14.6 &  63.7 & 17.2 \\
NUMERAL:CASE     &      66.7  & 40.0   &  58.8 \\
NUMERAL:INFL    &     75.0  &   100.0  &   78.9 \\
ORTH       &       11.3 &  68.9 &  13.5 \\
OTHER      &      5.3 &  17.5  & 6.1 \\
 {\color{BrickRed} PREP } &  {\color{BrickRed} 22.3 } &  {\color{BrickRed} 41.5 } &  {\color{BrickRed}  24.6 } \\
PRON       &      16.0 & 32.8  &  17.8 \\
 {\color{BrickRed} PUNCT } &  {\color{BrickRed} 14.6 } &  {\color{BrickRed} 58.0 } &  {\color{BrickRed} 17.2 } \\
SPELL      &    73.5 &  78.8  &  74.5 \\
 {\color{BrickRed} VERB } &  {\color{BrickRed} 15.6 } &  {\color{BrickRed}  35.1 }  &  {\color{BrickRed} 17.6 }\\
VERB:ASPECT   &    22.2  & 27.2 & 23.0 \\
VERB:FORM   &       40.0   &  66.7  & 43.5 \\
VERB:GEN    &    45.0  &   69.2 &  48.4 \\
VERB:INFL    &       48.6 &  90.0   & 53.6 \\
VERB:NUM     &   46.4 &  65.3 &  49.2 \\
 {\color{BrickRed} VERB:TENSE } &  {\color{BrickRed} 12.8 } &  {\color{BrickRed} 27.8 }  &  {\color{BrickRed} 14.4} \\
VERB:VOICE    &        100.0   &   50.0  & 83.3 \\
 {\color{BrickRed} WO } &  {\color{BrickRed} 4.8 } &  {\color{BrickRed} 54.6} &  {\color{BrickRed}  5.8} \\
\hline
  \end{tabular}}
  \caption{Error-type performance of GPT-3.5 in zero-shot setting for RULEC-GEC test set (Russian), measured using ERRANT.}
  \label{table:types2}
  \end{center}
\end{table}

\begin{table*}
  \begin{center}
  \scalebox{0.98}{
  \begin{tabular}{l|l|l|c|c|c}
\hline
      \textbf{Lang.} & \textbf{LM} & \textbf{Dataset} & $p(Trg) > p(Src)$ & $p(Hyp) > p(Trg)$ & $S(Hyp) > S(Trg)$ \\
\hline
      CZ & GPT-2 & AKCES-GEC & 71.9 & 25.6 & 12.2 \\
      DE & GPT-2 & Falco-MERLIN & 91.9 & 33.6 & 18.4 \\
      EN & GPT-2 (xl) & BEA-19 (dev.) & 90.3 & \textbf{54.7} & \textbf{52.5} \\
      & & CoNLL-14 & 86.0 & \textbf{57.2} & \textbf{50.6} \\
      ES & GPT-2 & COWS-L2H & 90.0 & 41.3 & 42.1 \\
      RU & GPT-3 & RULEC-GEC & 90.5 & \textbf{50.2} & \textbf{50.1} \\
      SV & GPT-Sw3 6.7B & SweLL-gold$^*$ & 93.1 & 30.7 & 27.4 \\
      UA & GPT-2 & UA-GEC & 68.3 & 35.3 & 26.2 \\
\hline
  \end{tabular}
  }
  \caption{Evaluation: how probabilities assigned to a sentence by pre-trained LM correlate with sentence type: \textit{Src} refers to original erroneous sentence; \textit{Trg} refers to gold-standard reference; \textit{Hyp} refers to correction suggested by GPT-3.5 in zero-shot setting.
  Reported numbers are percentages. \textit{S} in last column denotes Scribendi score.}
  \label{table:lm-critic}
  \end{center}
\end{table*}

As the pre-trained LMs, we use GPT-2 models for \href{https://huggingface.co/MU-NLPC/CzeGPT-2}{Czech}, \href{https://huggingface.co/dbmdz/german-gpt2}{German}, \href{https://huggingface.co/DeepESP/gpt2-spanish}{Spanish}, and \href{https://huggingface.co/Tereveni-AI/gpt2-124M-uk-fiction}{Ukrainian}.
We use \href{https://huggingface.co/gpt2-xl}{GPT-2 xl} for English, \href{https://huggingface.co/ai-forever/rugpt3large_based_on_gpt2}{GPT-3 large} for Russian, and \href{https://huggingface.co/AI-Sweden-Models/gpt-sw3-6.7b}{GPT-Sw3} with 6.7B parameters for Swedish.
All models are openly available on HuggingFace.
To evaluate GPT-3.5's quality, we aimed to use other models, as relying on GPT-3.5's own probabilities for its evaluation would be self-referential.
We chose monolingual models for this purpose, given their exposure only to language-specific data.

First, we collected pairs of erroneous source sentences (\textit{Src}) and their gold-standard references (\textit{Trg}) from multiple datasets.  We verified that the pre-trained LMs assign higher probabilities to the gold-standard references than the source sentences; see Table~\ref{table:lm-critic}, column $p(Trg)>p(Src)$ shows the percentage of instances in the dataset where \textit{Trg} LM's probability is higher than \textit{Src} LM's probability.  As we observe, grammaticality correlates with the LM's probabilities across all languages: LM's probabilities are lower for erroneous source sentences.

Next, we compare probabilities assigned to GPT-3.5's corrections (\textit{Hyp}) with the probabilities of the gold-standard references; see column $p(Hyp)>p(Trg)$ in Table~\ref{table:lm-critic}.
GPT-3.5 suggestions received higher LM's probabilities than the gold references only for English (over 54\%) and Russian (50.2\%).  For the other five languages, the GPT-3.5 corrections have lower LM's probabilities, which means a potentially lower level of grammaticality.

\vspace{4pt}
{\bf Evaluation with Scribendi Score:}
Another reference-less metric that we used is the Scribendi score~\cite{islam-magnani-2021-end}. It incorporates perplexity scores, a token sort ratio, and the Levenshtein distance score, and has been found to strongly correlate with human rankings.
Since the Scribendi score is language-independent, we can use it for each of the tested languages.

Perplexity indicates whether the suggested change improves the grammaticality and fluency of the sentence; it is calculated using the same LMs mentioned above.  The token sort ratio and the Levenshtein distance between the source and the prediction are incorporated into the formula to ensure that the overall meaning of the sentence remains unchanged and to penalize predictions with repetitions.

The rightmost column of Table~\ref{table:lm-critic} displays the percentage of GPT-3.5's suggestions that have higher Scribendi scores than the gold references. The results align with the evaluation using LMs: for English and Russian slightly over 50\% of the hypotheses proposed by GPT-3.5 are ranked higher. As observed, there is a correlation between the two reference-less metrics, although the Scribendi score penalizes predictions that significantly rephrase the source sentences. %Both manual and automatic evaluation underline the need for improved evaluation metrics for GEC, which also could potentially be used in combination.

\begin{figure*}[t]
\begin{center}
  \includegraphics[scale=0.41]{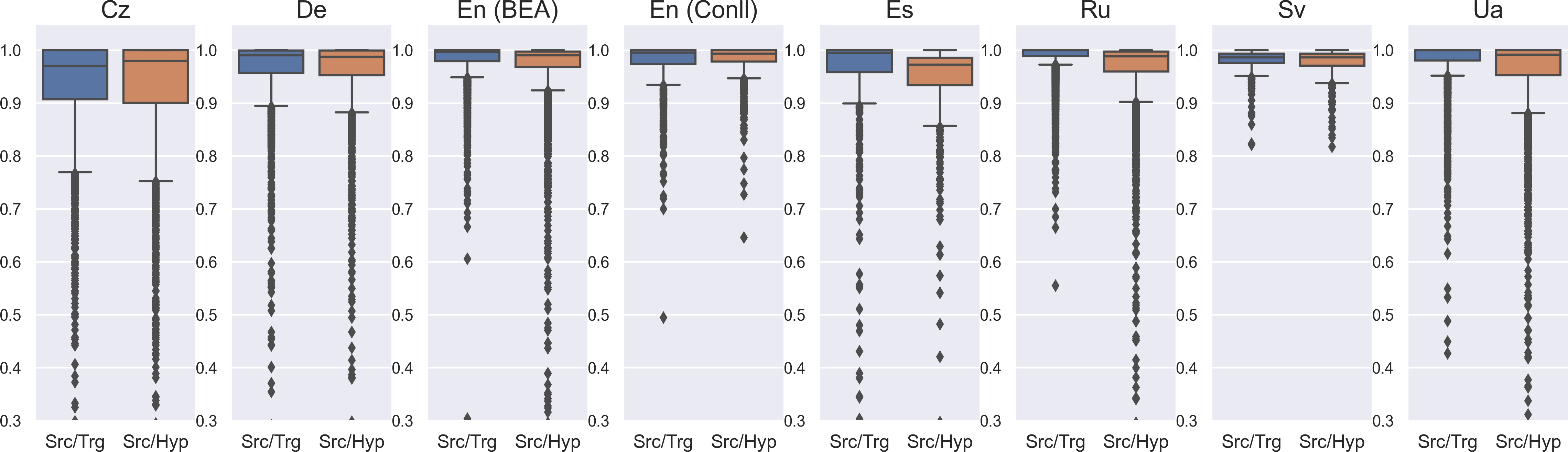}
  \caption{Cosine similarities between the source sentence embeddings and the target sentence embeddings (\textit{Src/Trg}) and cosine similarities between the source sentence embeddings and sentence embeddings of the hypotheses suggested by GPT-3.5 (\textit{Src/Hyp}).}
  \label{fig:similarity}
\end{center}
\end{figure*}

\begin{figure}[t]
\begin{center}
  \includegraphics[scale=0.46]{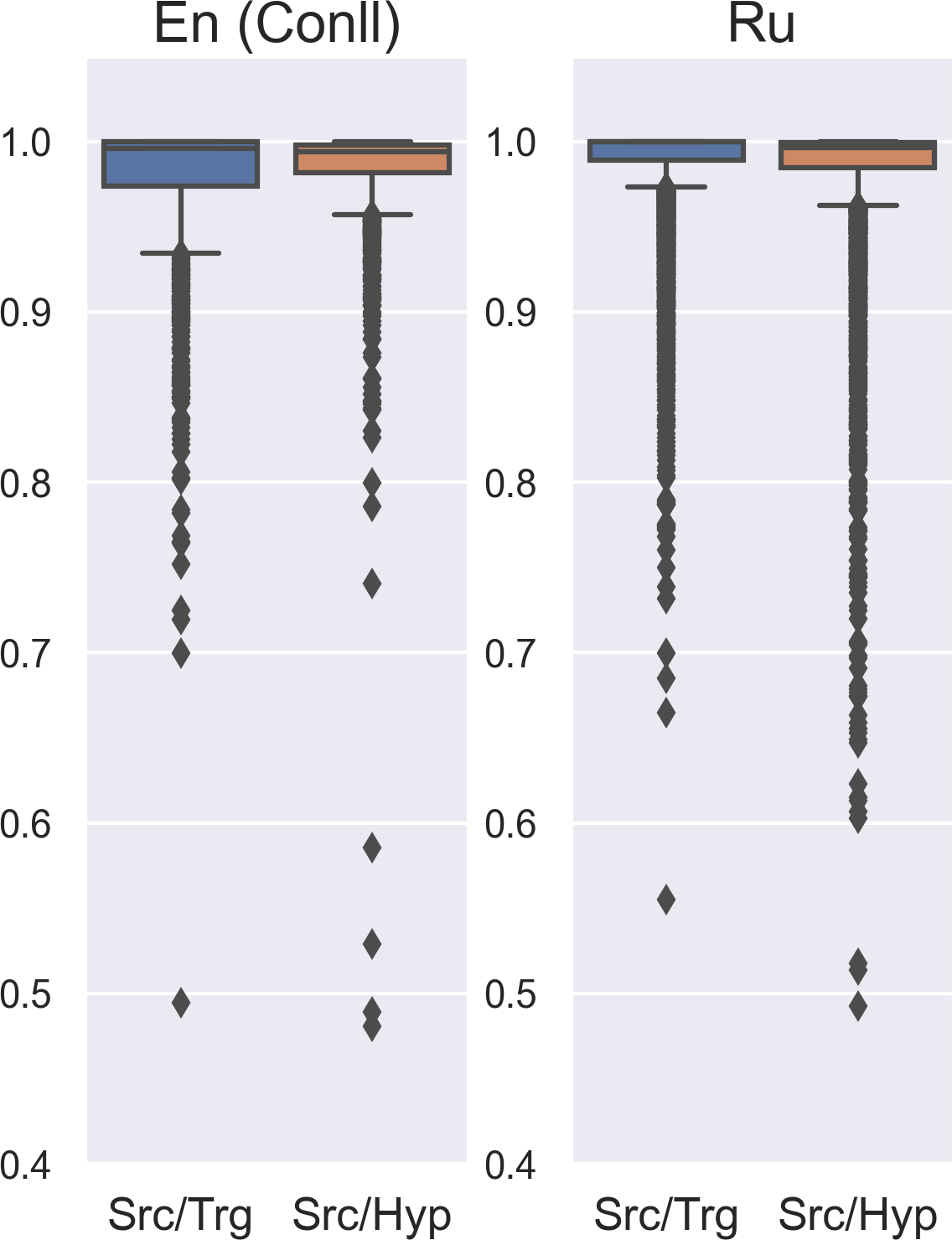}
  \caption{Cosine similarities between source sentence embeddings and target sentence embeddings (\textit{Src/Trg}) and between the source sentence embeddings and sentence embeddings of the re-ranked hypotheses suggested by other GEC models (\textit{Src/Hyp}).}
  \label{fig:similarity-re}
\end{center}
\end{figure}

\vspace{4pt}
{\bf Semantic Similarity:}
For additional automatic evaluation of the changes proposed by GPT-3.5 during grammatical error correction, we compared the embeddings of the source sentences, the reference (or target) sentences provided by human annotators, and the corrections (or hypotheses) suggested by GPT-3.5.  For this purpose, we utilize the Sentence Transformer model,\footnote{https://www.sbert.net/} a modification of the pretrained BERT that uses siamese and triplet network structures to derive semantically meaningful sentence embeddings~\cite{reimers-gurevych-2019-sentence}. For all languages, we use a multilingual model\footnote{distiluse-base-multilingual-cased-v2} to encode sentences~\cite{reimers-2020-multilingual-sentence-bert}. After that, we measured the cosine similarity (1) between the source embeddings and the target embeddings (\textit{Src/Trg}), and (2) between the source embeddings and the hypotheses embeddings (\textit{Src/Hyp}), to assess the semantic similarity of GPT-3.5's suggestions to the original input sentences.

From Figure~\ref{fig:similarity}, we observe that the difference between two cosine similarities (\textit{Src/Trg} and \textit{Src/Hyp}) varies across languages.  In the case of Czech and German, both the gold references and GPT-3.5 correction hypotheses exhibit significant differences from the source sentences.  This similarity in the deviations from the source sentences, observed in both the gold references and the model's suggestions, implies extensive corrections made by humans and by the model.  Although GPT-3.5 is known for occasionally altering the meaning of the source sentences, its corrections for English can be semantically even closer to the source sentence than the gold references, e.g., see the results for CoNLL.  For Russian, Spanish, and Ukrainian, the plots indicate that changes proposed by the model tend to transform the sentence meaning the most.

Compare the cosine similarities for CoNLL and for Russian RULEC in Figure~\ref{fig:similarity} with the analogous plots in Figure~\ref{fig:similarity-re}, where \textit{Hyp} denotes corrections generated by two other GEC models and re-ranked by GPT-3.5.  We can see that the semantics of the corrected sentences remain close to the semantics of the source sentences.  This is consistent with the observation that selecting top-ranked hypotheses results in higher precision, despite yielding a lower recall when compared to a zero-shot setting (see \texttt{rerank} and \texttt{zero-shot} in Table~\ref{table:zero}).

Cosine similarities between sentence embeddings, combined with LM scores and the Scribendi scores, as well as evaluation of GEC performance using reference-based metrics, suggests that GPT-3.5 may not be the optimal GEC model for Czech, German, Spanish, and Ukrainian.  For these languages, the overall GEC performance of GPT-3.5 is lower: LMs scores show that only a small percentage of suggested corrections are less ``surprising'' to language models than the gold reference sentences; the Scribendi scores are also modest, which means that sentences lexically deviate from the input and might have repetitions.

For Russian, the results indicate that GPT-3.5 can correct more errors than any other model (see the highest recall for RULEC in Table~\ref{table:zero}), but this comes at the expense of reduced semantic similarity between the source sentences and the proposed corrections.  Performance for English is the most encouraging: although precision is lower compared to other GEC models, recall is higher, and the semantic similarity between the source sentences and suggested corrections is notably high.  Re-ranking achieves even higher similarity while maintaining high recall, since GPT-3.5 prioritizes fluency.  Performance for Swedish is also promising, though we cannot compare it with other GEC models.  Corrections suggested by GPT-3.5 stay semantically similar to the source sentences, which may also result from under-corrected errors---this will agree with lower LM and Scribendi scores assigned to the correction hypotheses.

These findings suggest that while GPT-3.5 can be used for GEC, its efficacy varies across languages.  For certain languages, it might offer corrections that alter the intended semantics of the sentence.

\section{Conclusions}
\label{conclusion}

In this work, we evaluate the performance of GPT-3.5 models for GEC across different languages and experimental settings.
To our knowledge, this is the first paper that evaluates GPT-3.5 for GEC across the number of languages that we cover.
The results indicate that GPT-3.5 has strong error-correction capabilities for various languages.
However, current reference-based evaluation methods tend to underestimate its performance, because GPT-3.5 generates fluent sentences, often not following the minimum-edit principle.
We perform several types of evaluation of the corrections suggested by GPT-3.5: manually and by using reference-less metrics focused on sentence fluency and semantic similarity to the original input.

We find that GPT-3.5's behavior varies by language.  For Czech, German, Russian, Spanish, and Ukrainian, it tends to modify the input more freely, not always achieving superior performance.
In contrast, for English, GPT-3.5 achieves high recall, generates fluent sentences, and does not significantly alter sentence semantics.
For Swedish, GPT-3.5 archives relatively high precision and generates corrections that do not deviate semantically from the input sentences, which may, though, result from under-corrected errors, therefore, it requires manual evaluation.

The results of our experiments show that using GPT-3.5 to re-rank correction hypotheses generated by other GEC models outperforms GPT-3.5's own performance in both the zero-shot and fine-tuning settings for English and Russian.  This indicates that GPT-3.5 can be employed in a setup with smaller fine-tuned GEC models.  Manual evaluation of the corrections suggested by GPT-3.5 for English and Russian shows that the model often struggles with correcting: punctuation, articles, preposition errors, and verb tense errors in English; and errors in punctuation, in government between words, and in lexical compatibility in Russian.

Based on our evaluation, we emphasize the need to develop new evaluation metrics for GEC.  Traditional reference-based metrics may not adequately capture the advancements of modern LLMs.  We propose that a new metric should be based on multiple gold-standard references from human annotators, and it should take into account to what extent the {\em meaning} of the corrected sentence deviates from the meaning of the input.

Considering that GPT-3.5 can be further improved for various error types (particularly for languages other than English) and the challenges in evaluating its correctness, we find that it has the potential for integration into GEC systems, such as an ensemble with other models, serving as a re-ranker of correction hypotheses.  In future work, we plan to broaden our studies to more languages, experiment with freely available large language models, and perform a thorough in-depth human evaluation for all languages under study.

\section*{Limitations}

This work presents preliminary experiments on GEC for several languages using GPT-3.5 models, which were started by other researchers.  Our experiments in a zero-shot setting are limited to using only two prompt messages for all languages, as experimenting with various prompts for different languages was not the focus of our work. Experiments with fine-tuning were done with only three languages, without changing any default parameters due to financial limitations.  In the future, we plan to utilize freely-available large language models.  As for re-ranking hypotheses generated by other GEC models, we had resources to train only Russian GEC model; English GEC model was available for public use.  We plan to extend re-ranking experiments with the other languages in future work. We also did not experiment with providing GPT-3.5 with any correction examples, as was done in~\cite{fang2023chatgpt}.

Our manual evaluation of generated corrections is limited: we were unable to involve more native annotators for English and Russian and to perform any manual evaluation for other languages. In future work, we hope to have more resources to invite more annotators. However, we find estimation of GEC performance problematic: reference-based methods do not allow to get a realistic evaluation, while manual evaluation is very expensive and time-consuming, especially if the number of languages is large.

We used only publicly available resources for all conducted experiments. Regarding GPT-3.5, despite the official API being available to any registered user, we found a number of difficulties in using it in research experiments: black-box models are constantly updated, making it hard to verify any results obtained with them; it is difficult to control the model's behavior in a zero-shot setting because it frequently does not follow the required format; experiments with fine-tuning can be expensive and not affordable to many researchers.

\section*{Ethics Statement}

We use only publicly available resources for all conducted experiments. All annotators were volunteer students who performed the tasks as a part of their studies and received credits for it.

\bibliographystyle{lrec-coling2024-natbib}
\bibliography{reference}

% \section{Language Resource References}
% \label{lr:ref}
% \bibliographystylelanguageresource{lrec-coling2024-natbib}
% \bibliographylanguageresource{languageresource}

\end{document}